\documentclass[runningheads]{llncs}
\usepackage{algorithm2e}
\usepackage{makecell}
\usepackage{url}
\usepackage{cite}
\usepackage{subfig}
\usepackage{amssymb}
\usepackage{amsmath}
\usepackage{graphicx}
\usepackage{hyperref}
\begin{document}
\title{Towards Efficient and Data Agnostic Image Classification Training Pipeline for Embedded Systems}
\titlerunning{Towards Efficient and Data Agnostic Image Classification}
\author{Kirill Prokofiev\orcidID{0000-0001-9619-0248} \and
Vladislav Sovrasov\orcidID{0000-0001-6525-2602}}
\authorrunning{K. Prokofiev et al.}
\institute{Intel
\email{kirill.prokofiev@intel.com}, \email{vladislav.sovrasov@intel.com}}
\maketitle              % typeset the header of the contribution
\begin{abstract}
  Nowadays deep learning-based methods have achieved a remarkable progress at the
  image classification task among a wide range of commonly used datasets (ImageNet, CIFAR,
  SVHN, Caltech 101, SUN397, etc.). SOTA performance on each of the mentioned datasets
  is obtained by careful tuning of the model architecture and training tricks according to
  the properties of the target data. Although this approach allows setting academic records,
  it is unrealistic that an average data scientist would have enough resources to build a sophisticated training pipeline
  for every image classification task he meets in practice. This work is focusing on reviewing
  the latest augmentation and regularization methods for the image classification and
  exploring ways to automatically choose some of the most important hyperparameters:
  total number of epochs, initial learning rate value and it's schedule.
  Having a training procedure equipped with a lightweight modern CNN architecture (like MobileNetV3 or EfficientNet),
  sufficient level of regularization and adaptive to data learning rate schedule,
  we can achieve a reasonable performance on a variety of downstream image classification
  tasks without manual tuning of parameters to each particular task.
  Resulting models are computationally efficient and can be deployed to CPU using the OpenVINO{\texttrademark} toolkit.
  Source code is available as a part of the OpenVINO{\texttrademark} Training
  Extensions\footnote{\url{https://github.com/openvinotoolkit/training_extensions}}.

\end{abstract}
\section{Introduction}

Throughout the past decade deep learning-based image classification methods have made
a great progress increasing their performance by 45\%\cite{zhai2021scaling} from the AlexNet\cite{alexnet}
level on well-known ImageNet benchmark \cite{imagenet_cvpr09}.
Although SOTA results are obtained by a fine-grained adaptation of all the training pipeline components to
the target task, in practice a data engineer could not have enough resources to do it.
Typically to solve an image classification task we need to make the following decisions:
\begin{itemize}
  \item Choose a model architecture;
  \item Build data augmentation pipeline;
  \item Choose optimization method and it's parameters (learning rate schedule, length of the training);
  \item Apply some extra regularization methods in case of overfitting;
  \item Apply additional techniques to handle hard classes
  imbalance or high level of label noise if needed.
\end{itemize}
Wrong decisions on each step can hurt the resulting classification model performance or bring misalignments
with the requirements to computational complexity, for instance.
At the same time, techniques and models which are successful on one dataset may not be
beneficial on others, i.e. they are not dataset-agnostic. From the practical perspective, although reaching the ultimate performance is
very demanding task, one can still think about some fail-safe training configuration
that would allow obtaining moderate results on a wide range of middle-sized image classification datasets
with minimum effort to further tuning. In this work we aim to propose such a configuration for a couple of
modern computationally effective architectures: EfficientNet-B0 \cite{Tan2019EfficientNetRM} and MobileNetV3\cite{Howard2019SearchingFM} family.
To reach the goal we added the adaptability to the optimization process of scheduling the weights,
designed a robust initial learning rate estimation heuristic and curated a set of
regularization techniques that suit well to the considered model architectures.

In brief, the key contributions of this paper can be summarized as follows:
\begin{itemize}
  \item Designed an optimization controlling policy that includes an optimal initial learning rate estimator and
  a modified version of the ReduceLROnPlateau \cite{reduceonplateau} scheduler equipped with an early stopping procedure;
  \item Proposed a way of applying Deep Mutual Learning \cite{Zhang2018DeepML} to reduce overconfidence of model predictions;
  \item For each of the considered models curated a suitable bundle of data augmentation
  and regularization methods and validated it on various middle-sized downstream image classification datasets.
\end{itemize}

\section{Related Work}
\paragraph{Optimal learning rate estimation and scheduling.} The problem of
initial learning rate setting can be seen as a general hyperparameter optimization and, thus,
a variety of general methods can be directly applied \cite{optuna_2019, liaw2018tune}.
From the other side, several simple heuristics were designed to directly tackle it \cite{Mukherjee2019ASD, fastai_lr}.
The last approach is more lightweight because it doesn't imply dependencies on any hyperparameters estimation frameworks,
but at the same time, it is not very reliable since assumptions that these algorithms are based on, could
not strictly hold in a wide range of real-world tasks. Considering a fine-grained grid search as the most robust method,
we believe that a combination of a hyperparameters tuning framework and properly designed trials execution process
will be a good trade-off between robustness and accuracy. When the initial learning rate is chosen, the schedule and
stopping criteria define the final result of the training.
Typically researchers set reasonable amount of training epochs in advance, especially if they
focus on a single dataset, but to save computational resources and avoid overfitting it is beneficial to
stop training early \cite{early_stop, Zhang2017UnderstandingDL}. Early stopping can break logic of popular
schedulers like cosine or 1cycle\cite{Smith2019SuperconvergenceVF},
especially if too large number of total epochs was set initially, so
in this work we use drops to decrease learning rate in combination with initial linear warm-up.

\paragraph{Data augmentation.}
Recently a wide range of effective data augmentation methods has been proposed \cite{Naveed2021SurveyIM, Harris2020FMixEM}.
Augmix \cite{hendrycks2020augmix} allows combining several
simple augmentations (like random crop, color perturbation, rotation, flip) into a pipeline
with adjustable applying policy. That allows to use Augmix as a replacement for classic sequential pipeline of transformations.
In case of high capacity models, additional mixing-sample augmentations like fmix \cite{Harris2020FMixEM} can be applied on the top of Augmix output
to further diversify the input data.

\paragraph{Regularization.}
To achieve higher classification accuracy, a training pipeline should keep a balance between
fitting ability and regularization. We have tested a lot of approaches
acting on different directions: continuous dropout \cite{Shen2018ContinuousD}, mutual learning \cite{Zhang2018DeepML},
label smoothing and confidence penalty \cite{Pereyra2017RegularizingNN}, no bias decay \cite{bagOfTricks},
and found a suitable combination for each of considered architecture type.
Complexity of data augmentation, batch size and learning rate values can also be viewed as regularization factors.
Taking this into account, transferring regularization parameters between different datasets and architectures
should be done with care.

\paragraph{Optimization.}
Currently even SOTA approaches in classification still use SGD
\cite{Zhai2021ScalingVT, Pham2020MetaPL} for finetuning on small target datasets,
while use stronger adaptive optimizers \cite{Loshchilov2019DecoupledWD, Zeiler2012ADADELTAAA}
for initial pre-training with huge amount of samples. We are aiming to finetune already pre-trained models
and choose the SGD-based Sharpness-Aware Minimizer (SAM) \cite{Foret2020SharpnessAwareMF} as a default optimizer.
SAM, like SGD, preserves the mentioned fitting-regularization tradeoff, it's authors claim that SAM also
provides robustness to label noise. This property is useful if we aim to build a reliable training pipeline.

\section{Method}
In this section, we describe the overall training pipeline from models architectures to training tricks.

\subsection{Models}
We chose MobileNetV3 \cite{Howard2019SearchingFM} and EfficientNet \cite{Tan2019EfficientNetRM}
as base architectures for performing image classification.
Namely, we conducted all the experiments on MobileNetV3 small 1x, large 0.75x, large 1x and EfficientNet-B0.
The chosen models form a strong performance accuracy trade-off in the range from 0.06 to 0.4 GFLOPs, which is
enough in most of edge-oriented applications.

\subsection{Training tricks}
\label{subsec:tricks}

\paragraph{Data Augmentation.}

Properly chosen data augmentation pipeline can boost classification accuracy on a variety of downstream tasks.
At the same time, optimal augmentations are different for different datasets. Thus, any hand-crafted pipeline
would be suboptimal and considering this, our goal is to to find a pipeline that maintains fitting-regularization tradeoff for
MobileNetV3 and EfficientNet.
After conducting experiments with modern techniques \cite{hendrycks2020augmix,Zhang2018mixupBE,Yun2019CutMixRS,Harris2020FMixEM}, we found
AugMix\cite{hendrycks2020augmix} with a pre-trained on ImageNet policy is the most beneficial for our setup. When we add
MixUp\cite{Zhang2018mixupBE}, CutMix\cite{Yun2019CutMixRS} or FMix\cite{Harris2020FMixEM} to the pipeline
we observe a performance drop compared to pure AugMix. This indicates that AugMix provides a better
fitting-regularization tradeoff for the chosen lightweight models, while MixUp-like augmentations
are too hard for them.

\paragraph{Optimization.}

Conventionally, SGD with momentum is widely used for fine-tuning in downstream classification tasks \cite{Zhai2021ScalingVT, Pham2020MetaPL}.
An extension to SGD, Sharpness Aware Minimization (SAM) \cite{Foret2020SharpnessAwareMF}, allows to achieve higher results in
the fine-tuning than SGD, while requiring two forward-backward passes of the model per training iteration. We operate with lightweight models,
so additional cost of SAM is not critical. We also tried AdamW \cite{Kingma2015AdamAM}, but it performed slightly worse than SGD.

Additionally, we employ no bias decay (turning off weight decay for biases in all layers) \cite{bagOfTricks} for better generalization.

\paragraph{Optimal Learning Rate Estimation.}

\begin{figure}[t]
  \centering
  \subfloat[fast-ai]
  {{\includegraphics[width=.5\textwidth]{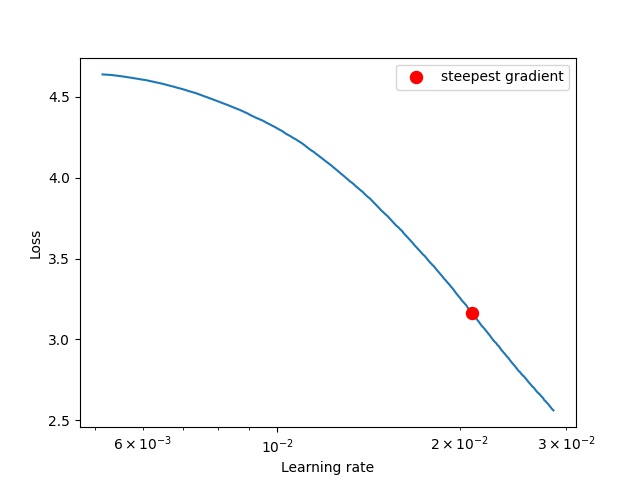}}
  }
  \subfloat[fast-ai with one epoch pretraining]
  {{\includegraphics[width=.5\textwidth]{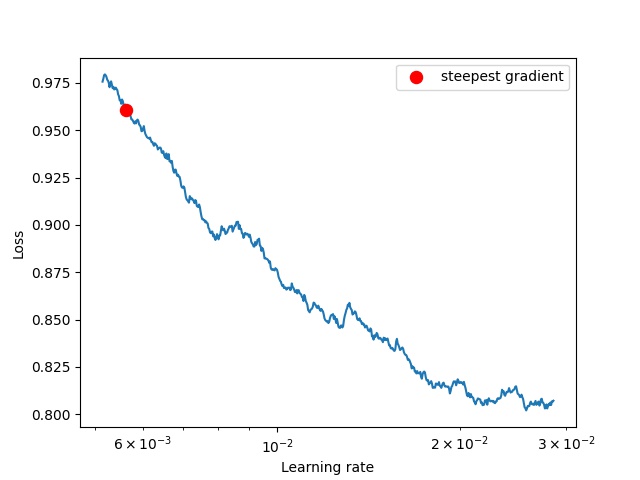}}}
  \caption{Without pretraining the estimated learning rate is too large ($\approx 0.02$). Whereas with one epoch pretraining the
   estimated learning rate is more suitable for the CIFAR-100 dataset ($\approx 0.0056$).}
  \label{fig:fast_ai}
\end{figure}

Since SGD performance on a given dataset is highly correlated with the initial learning rate magnitude, we have to
incorporate estimation of this parameter into our training pipeline. Straightforward approach is to use grid or random search
within a given range, but we were focusing on less time consuming methods. Fast-ai's heuristic \cite{fastai} can generate
learning rate proposal after performing several training iterations, but in our experiments it tends to output too high
values which destroy ImageNet initialization even if warmup strategy is applied. To overcome this problem, we propose to finetune
the model with a pre-defined small learning rate for one epoch on the target data and then run the original fast-ai algorithm.
In this case the model would react to fast increase of the learning rate more smoothly and the algorithm will select a value
which is quite close to one located with the grid search (see Figure~\ref{fig:fast_ai}). Fast-ai's heuristic with pre-training is
almost as lightweight as the original one: for instance if the training is scheduled for 100 epochs,
one extra epoch for finetuning will introduce only 1\% of additional overhead.

For fine-grained learning rate selections we use Tree-structured Parzen Estimator (TPE) from the Optuna \cite{optuna_2019} framework.
We set optimization criterion as top-1 accuracy on the validation subset after training on the target data for several epochs.
We use median trial pruning criterion which breaks a trial if the best intermediate result is worse than the median of intermediate results of
the previous trials at the same step. Pruning heuristic also could be used for grid search, but TPE also generates locations of the
next trials based on previous trials locations instead of random choice or using a grid, which allows us to set a lower total trails limit compared to the
grid search. We set TPE as a default initial learning estimator in our pipeline, though fast-ai with pre-training could be used
in case of limited resources.

\paragraph{Learning Rate Scheduling and Early Stopping.}

Complexity of the input dataset can drastically vary and thus the number of epochs that are sufficient for training also varies.
For efficient training we have to adapt to different data. To achieve this, we need a flexible learning rate schedule and
reliable early stopping criterion. Popular SOTA schedulers like cosine \cite{Loshchilov2017SGDRSG} and 1cycle\cite{Smith2019SuperconvergenceVF}
require a pre-defined number of epochs and the shape of the produced learning rate curve depends on it. If we, for an instance,
predefine 200 training epochs and early stopping criterion returns a stop flag on the 20th epoch, the training will be stopped at an unstable
state of the model, because convergence hasn't been reached yet. If we take into account this drawback and prohibit early stopping criterion
returning a stop flag till the half of the training pass, this will cause too long training, but the model will converge.

To overcome these problems we propose to use a modified version of ReduceLROnPlateau\cite{reduceonplateau} scheduler denoted hereafter as ReduceLROnPlateauV2.
In ReduceLROnPlateauV2 we force learning rate decay if 75\% of the maximum training length had been reached,
but learning rate drop have not been performed yet.
This helps model to converge when average training loss, which we use as a criterion for learning rate drop, is unstable.
Also we incorporated early stopping into ReduceLROnPlateauV2: if the learning rate was decayed to a pre-defined minimal value and the best
top-1 score on validation subset hadn't been improving for a predefined number of epochs, we stop the training.

Additionally we use 5 epoch linear learning rate warmup and, following no bias decay practice, we increase
learning rate for biases by a factor of 2.

\paragraph{Mutual Learning.}

\begin{figure}[t]
  \centering
  {\includegraphics[width=0.9\textwidth]{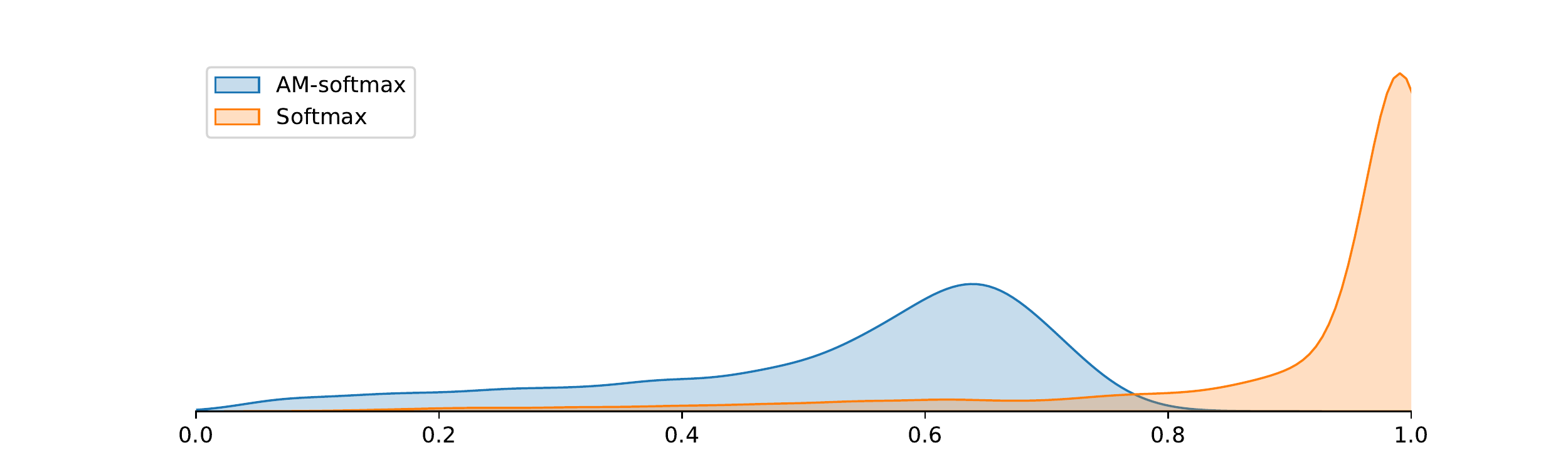}}
  \caption{Distributions of the most probable class confidence produced by the fast model on validation data. If
  slow student is trained with AM-Softmax, fast student does not tend to return overconfident predictions. Experiment is
  conducted with a pair of MobileNetV3 small models on Cars Dataset \cite{KrauseStarkDengFei-Fei_3DRR2013} containing 196 classes.}
  \label{fig:dist}
\end{figure}

To further boost the performance of classification models we tried to apply deep mutual learning (DML) \cite{Zhang2018DeepML}.
This technique implies mutual learning of a collection of simple student models. As a result, performance of each student
individually is supposed to be increased. We directly apply this technique to pairs of identical models and didn't
obtain substantial accuracy gain. This framework could also be used in a different manner: if one of the students
has a stronger regularization and trains slower than the others, it may prevent others from overfitting and
softly transfer properties of it's regularized distribution to faster students. We applied the second setup
to pairs of models with the same architecture, but different loss functions. Fast student is trained with the
cross-entropy loss while slow one is trained with the AM-Softmax\cite{AMSoftmax} loss with scale $s=1$ and margin $m=0$.
Such settings of the angular loss make output distribution of the slower model less confident and it also pushes
the faster model to output a smoother distribution while strong discriminative properties formed
by AM-Softmax are transferred to the faster model as well.
In the described scheme the overall training losses are defined as follows:

\begin{displaymath}
  \label{eq:equationDML}
  \begin{split}
    L_{fast}(x) = L_{CE}(p_1(x), y) + D_{KL}(p_2(x) || p_1(x))\\
    L_{slow}(x) = L_{AMS}(p_2(x), y) + D_{KL}(p_1(x) || p_2(x))
  \end{split}
\end{displaymath}
where $(x,y)$ is a training sample and it's label, $D_{KL}(p||q)$ is Kullback Leibler divergence between discrete distributions $p$ and $q$,
$p_1(x)$ -- distribution estimated by the fast model on the sample $x$,  $p_2(x)$ -- distribution estimated by the slow model on the sample $x$.
Results of applying this approach are demonstrated on the Figure~\ref{fig:dist}.

Drawback of the DML is significant increase of the training time and memory footprint. Taking this into account, we decide to include
this technique only to the training strategy for MobileNetV3 family, besides for EfficientNet-B0 it is not so beneficial.
Instead, we train EfficientNet-B0 with AM-Softmax directly. In that case  $s = \max(\sqrt{2} \cdot \log(C - 1), 3)$
(where $C$ is the number of classes) and margin $m=0.35$.

\section{Experiments}
This section describes the evaluation process, metrics, datasets and reports the evaluation results. For a given model family we use the
same training techniques and parameters across all the considered datasets;
this allows us to validate how well the proposed training pipeline can work on variable data.

\subsection{Datasets}
To validate the proposed approach we choose 11 widespread classification datasets, that are listed
in Table \ref{tab:data}.

\begin{table}
  \caption{Image classification datasets which were used for training.}
  \label{tab:data}
  \centering
  \begin{tabular}{l|c|c|c}
    Dataset & Number of classes & \multicolumn{2}{c}{Number of Images} \\
            &                   & Train     &   Validation \\ \hline
    Oxford-IIIT Pets\cite{parkhi12a} & 37 & 3680 & 3369 \\
    Describable Textures (DTD)$^*$ \cite{cimpoi14describing} & 47 & 4826 & 814 \\
    Oxford 102 Flowers$^*$ \cite{Nilsback08} & 102 & 6614 & 1575 \\
    Caltech 101$^*$ \cite{FeiFei2004LearningGV} & 101 & 6941 & 1736 \\
    Cars Dataset \cite{KrauseStarkDengFei-Fei_3DRR2013} & 196 & 8144 & 8041 \\
    Birdsnap$^*$ \cite{birdsnap} & 500 & 47386 & 2443 \\
    CIFAR-100 \cite{Krizhevsky2009LearningML} & 100 & 50000 & 10000 \\
    Fashion-MNIST \cite{fashionmnist} & 10 & 60000 & 10000 \\
    SVHN \cite{Netzer2011ReadingDI} & 10 & 73257 & 26032 \\
    Food-101 \cite{bossard14} & 101 & 75750 & 25250 \\
    SUN397$^*$ \cite{Xiao2010SUNDL} & 397 & 92440 & 16314 \\
    \hline
    \multicolumn{4}{l}{\footnotesize{$^*$ for experiments on these datasets we do custom random splits.}}
  \end{tabular}
\end{table}

\subsection{Evaluation protocol}
To evaluate image classification models besides commonly used \textit{top-k} accuracy
metric, we also calculate mean average precision (mAP) in the same sense as it is
considered in person re-identification field \cite{Zheng2015ScalablePR}. We would denote a set of $C$ classes as
a gallery and each of $N$ samples in the evaluation set as a query. Then, in terms of
the retrieval task, we have to rank the gallery by similarity with the query, compute the average precision
of this query, and then collect the mAP. In terms of the classification task, the similarity is the
predicted probability of a class, and, thus, computing average precision is straightforward:
\begin{displaymath}
  mAP=\frac{\sum_{q=1}^Q AP(q)}{Q}=\frac{\sum_{i=1}^N AP(i)}{N}=\frac{1}{N}\sum_{i=1}^N \frac{1}{K_i},
\end{displaymath}
where $K_i$ is the index of the probability of ground truth class in the sorted set
of predicted class probabilities for a sample. $AP(i)$ varies from 1 (if the model prediction is always correct)
to $\frac{1}{C}$ (if the model always assigns the lowest probability to the true class). mAP can
be viewed as an aggregation of all $C$ possible \textit{top-k} scores and reveals the ranking
ability of the evaluated classification models.

After single dataset evaluation metrics are defined, we also have to define a way to compare
different strategies for a given model taking into account results on a set of datasets. In this work,
we will follow \cite{Zhai2019TheVT} and consider metrics averaging across datasets as a measure of the quality
of a training strategy for a given model. This approach doesn't imply any weighting procedure,
so each dataset equally contributes to the final metric.

\subsection{Results}

We trained all the models starting from the ImageNet-pretrained weights.
We consider the following training setup as a baseline: training length is 200 epochs;
learning rate is annealed to zero with cosine schedule; optimization is performed with SGD with momentum;
training images are augmented with random flip, random color jittering, random crop and random rotation;
dropout with $p=0.2$ is applied to the classifier layer; learning rate is set as
an average of optimal learning rates obtained by grid search over all the 11 considered datasets for each model
individually.

\begin{table}[h]
  \caption{Results of our adaptive training pipeline against baseline.}
  \label{tab:final_avg_results}
    \centering
    \begin{tabular}{l|c|c|c|c}
      Model & AVG \textit{top-1}, \% & AVG \textit{top-5}, \% & AVG \textit{mAP} & AVG \textit{epochs} \\ \hline
      MobileNetV3-small baseline & 82.24	& 95.29 & 82.35 & 200\\
      Our MobileNetV3-small & 85.00 & 96.15 & 88.01 & 86 \\ \hline
      MobileNetV3-large-0.75x baseline & 85.28 & 96.30 & 87.03 & 200\\
      Our MobileNetV3-large-0.75x & 87.60 & 96.97 & 91.14 & 83 \\ \hline
      MobileNetV3-large baseline & 85.79 & 96.47 & 87.99 & 200 \\
      Our MobileNetV3-large & 88.26 & 97.34 & 91.79 & 81 \\ \hline
      EfficientNet-B0 baseline & 86.47 & 96.77 & 89.27 & 200 \\
      Our EfficientNet-B0  & 89.13 & 97.79 & 92.75 & 82\\
      \hline
    \end{tabular}
  \end{table}

For our adaptive pipeline we set the maximum training length to 200 epochs, maximum amount of trials in TPE
to 15 (each trial takes 6 epochs or less); learning rate search range for EffficientNet-B0 is $[0.001-0.01]$,
while for MobileNetV3 the range is $[0.005-0.03]$; Augmix with a pre-trained on ImageNet policy, random flip,
random crop and random rotation are employed for data augmentation; early stopping and learning rate decay patiences
are equal to 5 epochs; $\rho$ in SAM is 0.05. Weight decay is always set to $5\cdot10^{-4}$. Input resolution
for all versions of the considered models is $224\times224$, no test time augmentation is applied.

The final results are presented for all our models in the Table~\ref{tab:final_avg_results}. Our adaptive training
strategy clearly outperforms baseline by $2-3\%$ AVG top-1 and $3-5\%$ AVG mAP. Also it reduces
average number of epochs required for training more than twice: from $200$ to $86$ or less.

Top-1 scores by datasets are presented in the Table~\ref{tab:comp_metrics} as well as comparison with other solutions.
Our MobileNetV3-small with adaptive training outperforms both our baseline and one of the best publicly available
repositories with advanced training tricks for MobileNetV3-small (it includes label smoothing \cite{Pereyra2017RegularizingNN},
Mixup \cite{Zhang2018mixupBE}, no bias decay \cite{bagOfTricks}, EMA decay and learning rate warmup). At the same time, two
non-adaptive strategies for MobileNetV3-small perform on par. EfficientNet-B0 with adaptive strategy demonstrates results similar to ResNet-50 from
VTAB \cite{Zhai2019TheVT}, which is trained with heavyweight search over hyperparameters (learning rate, schedule, optimizers, batch size,
train preprocessing functions, evaluation pre-processing, and weight decay).
The results of heavyweight SOTA models are also presented in the Table~\ref{tab:comp_metrics} for the reference.

\begin{table}[h]
  \caption{Detailed comparison with other methods. Top-1 metric is presented.}
  \label{tab:comp_metrics}
    \centering
    \begin{tabular}{l|c|c|c|c|c|c|c|c|c|c|c} Model & \rotatebox[origin=c]{90}{CIFAR-100$^*$} &	\rotatebox[origin=c]{90}{DTD$^*$}	&	\rotatebox[origin=c]{90}{Food-101$^*$} &	\rotatebox[origin=c]{90}{SUN397} &	\rotatebox[origin=c]{90}{SVHN$^*$} &	\rotatebox[origin=c]{90}{Birdsnap} &	\rotatebox[origin=c]{90}{Caltech101} &
      \rotatebox[origin=c]{90}{Cars$^*$} &	\rotatebox[origin=c]{90}{Fashion-MNIST$^*$} &	\rotatebox[origin=c]{90}{Flowers$^*$}	&	\rotatebox[origin=c]{90}{Pets$^*$} \\ \hline
      Our MNV3-small baseline & 79.57 & 68.58 & 74.71 & 57.38 &	96.02 &	77.12	& 90.41 &	87.24	 & 95.13 &  93.89 &	84.65 \\
      Our MNV3-small & 83.49	&	72.45 & 79.59	& 62.83	& 97.29	& 79.04	& 92.94	& 91.33	& 95.74	& 94.86 & 85.55 \\
      MNV3-small$^+$  & 82.43 & 68.13 &	69.45 &	61.13	& 96.49	& 78.75 &	91.74	& 86.66	& 95.48	& 92.81 & 85.41 \\ \hline \hline
      MNV3-large-0.75x baseline & 81.05 & 72.28 & 80.97 & 61.97 & 96.13	& 81.39	& 92.87	& 91.18	& 95.02	& 95.36 & 89.86 \\
      Our MNV3-large-0.75x & 85.36	& 74.50 & 85.53 & 67.02 & 97.54 & 83.21	& 95.23	& 93.75	& 95.83	& 96.22 & 90.51 \\ \hline \hline
      Our MNV3-large baseline & 81.70 & 73.53 & 81.57	& 62.65	& 96.18	& 82.21	& 93.46	& 91.29	& 95.31	& 95.69 & 90.13 \\
      Our MNV3-large    & 86.24	& 76.49 & 85.77	& 67.96	& 97.57	& 84.1 & 95.42	& 93.76	& 96.17	& 96.63 & 91.21 \\ \hline \hline
      Our EffNet-B0 baseline & 84.84	& 74.81 & 83.75	& 64.43	& 96.88	& 80.03	& 94.54	& 90.46	& 95.80 & 95.26 & 91.74 \\
      Our EffNet-B0      & \textbf{86.52}	& \textbf{77.18} & 86.06	& 72.10 & \textbf{97.82} & 83.33 & 95.63 & 93.77 & 96.28 & 97.10 & 92.00 \\
      VTAB ResNet-50 \cite{Zhai2019TheVT} & 84.00 & 76.8 & - & - & 97.40 & - & - & - & - & \textbf{97.40} & \textbf{92.6} \\ \hline \hline
      % Shake-Shake+AutoAugm \cite{Cubuk2019AutoAugmentLA} & 85.70 & - &  - & - & 99.00 & - & - & - & - & - & - \\
      Inception v4 \cite{Kornblith2019DoBI} & 87.5 & 78.1 & 90.00 & - & - & - & - & 93.30 & - & 98.50 & 94.50 \\
      EffNet-B7+SAM \cite{Foret2020SharpnessAwareMF} & 92.56 & - & 92.98 & - & - & - & - & 94.82 & - & 99.37 & 96.03 \\
      \hline
      \multicolumn{12}{l}{\footnotesize{$^*$ for these datasets we do splits as defined by the authors, for others we use custom splits}} \\
      \multicolumn{12}{l}{\makecell[l]{\footnotesize{$^+$Implementation is taken from \url{https://github.com/ShowLo/MobileNetV3}, learning rate is set}\\ \footnotesize{the same as for our baseline.}}}
    \end{tabular}
  \end{table}

\subsection{Ablation study}
We conduct an ablation study by removing each single component of our training pipeline, while others are enabled (see Table~\ref{tab:ablation}).
For experiments we use all the data except the SUN397 for MobilenetV3 and subset of 6 datasets for EfficientNet-B0 (CIFAR-100, DTD, Flowers, Cars, Pets, Caltech101).
Each of the training tricks has roughly equal gain ($<2\%$ top-1), excepting AM-Softmax related ones that add $2-3\%$ of mAP to the result.
\begin{table}
  \caption{Impact of each training trick to MobileNetV3-large and \mbox{EfficientNet-B0} results.}
  \label{tab:ablation}
  \centering
  \begin{tabular}{l|c|c|c|c}
    Configuration & \multicolumn{2}{c|}{MobileNetV3-large} & \multicolumn{2}{c}{EfficientNet-B0}\\
    & AVG \textit{top-1} & AVG \textit{mAP} & AVG \textit{top-1} & AVG \textit{mAP} \\
    \hline
    Final solution & \textbf{90.63} & \textbf{94.17} & \textbf{90.63} & \textbf{94.14} \\
    w/o SAM & 89.74 & 93.34 & 90.44 & 94.14 \\
    w/o Mutual Learning & 89.58 & 90.71 & - & - \\
    w/o AugMix & 90.14 & 93.75 & 90.41 & 93.88 \\
    w/o NBD  & 90.16 & 93.49 & 90.43 & 94.14 \\
    w/o AM-Softmax & - & - & 90.51 & 92.14 \\
    w/o adaptive learning rate strategy & 89.97 & 93.38 & 90.16 & 93.83 \\
    \hline
  \end{tabular}
\end{table}

\section{Conclusion}
In this work, we presented adaptive training strategies for several lightweight image classification models that
can perform better on a wide range of downstream datasets in finetuning from ImageNet scenario
than conventional training pipelines if there are no resources available for heavyweight models and extensive hyperparameters optimization.
These strategies are featured with optimal learning rate estimation and early stopping criterion, allowing
them to adapt to the input datasets to some extent. The ability to adapt is shown by conducting experiments on 11 diverse image classification datasets.
\bibliographystyle{splncs04}
\bibliography{egbib}
\setcounter{section}{1}
\section*{Supplementary Material}

\subsection{Classification Tasks}
We provide a brief description of each task:

\begin{itemize}
  \item \textbf{Caltech101.} The task consists in classifying pictures of objects (101 classes plus a background clutter
  class), including animals, airplanes, chairs, or scissors. The image size varies, but it typically ranges from 200-300
  pixels per edge.
  \item \textbf{Oxford-IIIT Pets.} The task consists in classifying pictures of cat and dog breeds (37 classes with around 200 images
  each), including Persian cat, Chihuahua dog, English Setter dog, or Bengal cat. Images dimensions are typically 200
  pixels or larger.
  \item \textbf{Describable Textures (DTD).} The task consists in classifying images of textural patterns (47 classes, with 120 training images
  each). Some of the textures are banded, bubbly, meshed, lined, or porous. The image size ranges between 300x300 and
  640x640 pixels.
  \item \textbf{Oxford 102 Flowers.} The task consists in classifying images of flowers present in the UK (102
  classes, with between 40 and 248 training images per class). Azalea, Californian Poppy, Sunflower, or Petunia are
  some examples. Each image dimension has at least 500 pixels.
  \item \textbf{SUN397.} The Sun397 task is a scenery benchmark with 397 classes and, at least, 100 images per class.
  Classes have a hierarchy structure, and include cathedral, staircase, shelter, river, or archipelago. The images are
  (colour) 200x200 pixels or larger.
  \item \textbf{SVHN.} This task consists in classifying images of Google’s street-view house numbers (10 classes,
  with more than 1000 training images each). The image size is 32x32 pixels.
  \item \textbf{Cars Dataset.} The task consists in classifying images of the cars,
  where each class has been split roughly in a 50-50 split. Classes are typically at the level of Make, Model, Year,
  e.g. 2012 Tesla Model S or 2012 BMW M3 coupe.
  \item \textbf{Birdsnap.} The task of the fine-grained visual categorization of birds. There are 500
  species of North American birds represented, labeled by species. There are between 69 and 100 images per species.
  \item \textbf{CIFAR-100.} The task consists in classifying natural images (100 classes, with 500 training images
  each). Some examples include apples, bottles, dinosaurs, and bicycles. The image size is 32x32.
  \item \textbf{Fashion-MNIST.} Fashion-MNIST is a dataset of Zalando's article images. Each example is a 28x28 grayscale image, associated with a label from 10 classes.
  \item \textbf{Food-101.} The data set of 101 food categories. For each class, 250 manually reviewed test images are provided as well as 750 training images.
  On purpose, the training images were not cleaned, and thus still contain some amount of noise.
  This comes mostly in the form of intense colors and sometimes wrong labels. All images have a maximum side length of 512 pixels.

\end{itemize}

\subsection{Performance Evaluation on CPU}

We evaluate the performance of our models in the OpenVINO\texttrademark 2021.3 Toolkit. Results are presented in the Table~\ref{tab:perf}.
Models provide a performance-accuracy trade-off in $0.4$ GFLOPs budget.

\begin{table}[h]
  \caption{Performance on the Intel\textregistered  Xeon\texttrademark Gold 6230 2.1GHz CPU in OpenVINO\texttrademark 2021.3 Toolkit.
           Batch size is set to 1, input resolution is $224\times224$, inference precision is FP32.}
  \label{tab:perf}
    \centering
    \begin{tabular}{l|c|c|c|c}
      Model & GFLOPs & Parameters, M & FPS & Latency, ms \\ \hline
      MobileNetV3-small-1x    & 0.06 & 1.61 & 467.9 & 2.05  \\
      MobileNetV3-large-0.75x & 0.15 & 2.83 & 365.22 & 2.65 \\
      MobileNetV3-large-1x    & 0.22 & 4.31 & 336.09 & 2.89\\
      EfficientNet-B0         & 0.4 & 4.11 & 322.75 & 3.01 \\
      \hline
    \end{tabular}
  \end{table}

  \subsection{Extended Ablation Study}

  For the extended ablation study we use all the data except the SUN397 for MobileNetV3 and subset of 6 datasets for
  EfficientNet-B0 (CIFAR-100, DTD, Flowers, Cars, Pets, Caltech101) if nothing else is said.

  \paragraph{Data augmentations.}

  \begin{table}[h]
    \caption{Augmentation influence for the MobileNetV3-large.}
    \label{tab:augmentations_mnv3}
    \centering
    \begin{tabular}{l|c|c}
      Augmentation & \multicolumn{2}{c}{AVG metrics} \\
                                    & \textit{top-1} & \textit{mAP} \\ \hline
      Basic augmentations & 88.89 & 90.33 \\
      FMix & 89.00 & 91.17 \\
      AugMix & \textbf{89.37} & 90.53 \\
      CutMix & 89.35 & \textbf{91.50} \\
      MixUp & 88.11 & 90.41 \\
      AugMix + CutMix & 89.15 & 90.98 \\
      AugMix + MixUp & 88.31 & 90.36 \\
      AugMix + Fmix & 89.15 & 90.86 \\
      \hline
    \end{tabular}
  \end{table}

  The Table~\ref{tab:augmentations_mnv3} and Table~\ref{tab:augmentations_effnet} show the
  results of applying augmentations for MobileNetV3-Large and EffficientNet-B0.
  As basic augmentations, we adopt random flip, rotate, scale, and color jitter.
  AugMix reaches the highest AVG top-1 score, while it's combination with other *mix augmentations slightly degrades the quality.

  \begin{table}[h]
    \caption{Augmentation influence for the EfficientNet-B0.}
    \label{tab:augmentations_effnet}
    \centering
    \begin{tabular}{l|c|c}
      Augmentation & \multicolumn{2}{c}{AVG metrics} \\
                                    & \textit{top-1} & \textit{mAP} \\ \hline
      AugMix & \textbf{90.52} & 92.14 \\
      FMix & 90.06 & 92.77 \\
      CutMix & 90.19 & \textbf{93.12} \\
      AugMix + CutMix & 90.06 & 92.79 \\
      AugMix + Fmix & 90.02 & 92.72 \\
      \hline
    \end{tabular}
  \end{table}

\paragraph{Optimization.}

  Table~\ref{tab:SAM} shows the results of adopting SAM as optimizer for EfficientNet-B0 and MobileNetV3-Large.
  We set hyperparameter $\rho= 0.05$ for all our models as it was proposed in \cite{Foret2020SharpnessAwareMF}.

  \begin{table}
    \caption{Results of training EfficientNet-B0 and MobileNetV3-large with SAM.}
    \label{tab:SAM}
    \centering
    \begin{tabular}{l|c|c|c|c}
      Optimizer & \multicolumn{2}{c}{AVG metrics} & \multicolumn{2}{|c}{CIFAR100}   \\
      & \textit{top-1} & \textit{mAP}  & \textit{top-1} & \textit{mAP}  \\
      \hline
      MNV3-large 1x + SGD & 89.74 & 93.34 & 85.12 & 90.39 \\
      MNV3-large 1x + SAM & 90.56 & 93.97 & 86.24 & 91.73 \\
      EffNet-B0 + SGD & 89.16 & 91.78 & 85.43 & 90.13 \\
      EffNet-B0 + SAM & 90.37 & 92.20 & 86.76 & 89.69 \\
      \hline
    \end{tabular}
  \end{table}

  From Table~\ref{tab:ablation} the overall NBD influence on the algorithm could be seen.
  As it was mentioned in Section~\ref{subsec:tricks}, in case of NBD we scale the learning
  rate for biases by a factor of 2 and use warmup for 5 epochs.
  In Table~\ref{tab:NBD_settings} the impact of these settings could be seen.

\begin{table}
    \caption{Results of applying NBD to MobileNetV3-small with different settings.}
    \label{tab:NBD_settings}
    \centering
    \begin{tabular}{l|c|c}
      Optimization method & \multicolumn{2}{c}{AVG metrics} \\
      & \textit{top-1} & \textit{mAP} \\
      \hline
      Raw NBD & 86.85 & 90.01 \\
      NBD + 2*lr & 86.89 & 90.15 \\
      NBD + 2*lr + warmup & \textbf{87.15} & \textbf{90.15} \\
      \hline
    \end{tabular}
\end{table}

\paragraph{Learning Rate Scheduling and Early Stopping.}

  To validate our ReduceLROnPlateauV2 scheduler we compare it against cosine schedule for 200 epochs.
  For all the cases learning rate was fixed to 0.013, SGD was set as an optimizer.
  In this experiment we use all of the 11 datasets for MobileNetV3 and the subset of 7 datasets
  (CIFAR-100, FOOD101, DTD, Flowers, Cars, Pets, Caltech101) for EffficientNet-B0.
  ReduceLROnPlateauV2 strategy clearly outperforms cosine schedule (see Table~\ref{tab:controll_policy}),
  while maintaining 2x and less training length on all the considered datasets. Thus, applying
  adaptive training controlling allows us to mitigate computational overhead introduced by SAM or mutual learning techniques.

  \begin{table}
    \caption{Results of applying adaptive training controlling policy.}
    \label{tab:controll_policy}
    \centering
    \begin{tabular}{l|c|c|c|c}
      Loss & \multicolumn{2}{c|}{AVG metrics} & AVG \# of epochs & Max \# of epochs \\
            & \textit{top-1} & \textit{mAP} & & \\ \hline
      MNV3-large + Cosine & 86.01 & 88.05 & 200 & 200 \\
      MNV3-large + ReduceLROnPlateauV2 & \textbf{86.89} &	\textbf{88.46} & \textbf{81} & \textbf{107} \\
      EffNet-B0 + Cosine & 88.39 & 91.35 & 200 & 200 \\
      EffNet-B0 + ReduceLROnPlateauV2 & \textbf{88.67} &	\textbf{91.35} & \textbf{82} & \textbf{95} \\
      \hline
    \end{tabular}
  \end{table}

  \paragraph{Optimal Learning Rate Estimation.}

  For each of the considered model families we conduct extensive experiments to find the optimal learning rate on each dataset.
  We set the searching space to $[0.001, 0.1]$ and use grid search with a uniform grid of 10 points. A trial in each grid point takes 9 epochs.

\begin{figure}
    \centering
    {{\includegraphics[width=\textwidth]{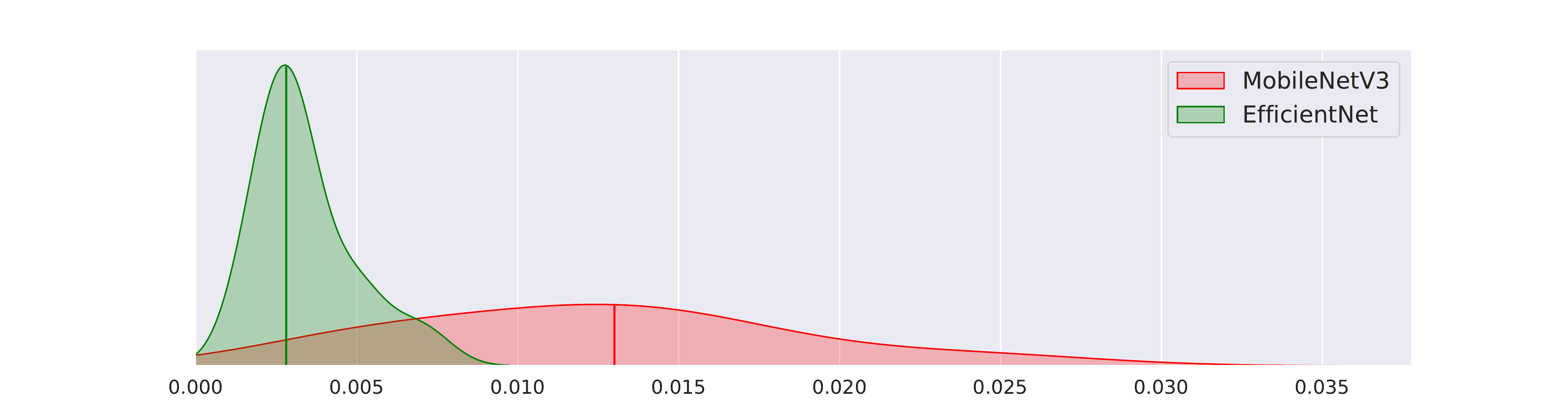}}}
    \caption{Distribution of optimal learning rates on the considered set of 11 classification tasks.}
    \label{fig:LR_ranges}
  \end{figure}

  We found that in average EfficientNet-B0 requires an order of magnitude lower learning rate than MobileNetV3 (see Figure~\ref{fig:LR_ranges}).
  According to our experiments, the most optimal search range for EfficientNet is $[0.001,0.01]$ with the average 0.003,
  while for MobileNetV3 the range is $[0.005,0.03]$ with the average 0.013.
  We use this information to limit boundaries for learning rate search algorithms. This helps to save time for learning rate estimation.

  For TPE we use discrete searching space with a step size 0.001.
  We use 6 epochs per trial and restrict the number of trials to 15. Trials are conducted on all the training data.
  This number of trials lets the algorithm to find a good value for learning rate in a reasonable amount of time.
  In the Table~\ref{tab:LR_estimation} we provide the results of all three methods for MobileNetV3-large.

  \begin{table}
    \caption{Learning rate estimation based on different algorithms.}
    \label{tab:LR_estimation}
    \centering
    \begin{tabular}{l|c|c|c}
        LR finder method & AVG \textit{top-1} & AVG \textit{mAP} & AVG time to estimate lr, min \\
        \hline
        Fast-ai & 88.27 & 90.37 & 2\\
        Grid search & 88.44 & 90.78 & 36\\
        TPE & 91.19 & 88.91 & 32\\
        \hline
      \end{tabular}
  \end{table}

  \paragraph{Mutual Learning.}
  We studied the impact of different DML settings to MobilenetV3-large (see Table~\ref{tab:DML}).
  Although after applying AM-Softmax to the slow student, classification accuracy increases by a small margin, mAP
  increases significantly. This indicates that the proposed setup of mutual learning indeed generates a model
  with more discriminative features and a greater ranking ability.

  \begin{table}
    \caption{Mutual learning with AM-Softmax validation results.}
    \label{tab:DML}
    \centering
    \begin{tabular}{l|l|c|c|c|c}
      Model & Method & \multicolumn{2}{c|}{AVG metrics}  & \multicolumn{2}{c}{Cars dataset}   \\
      && \textit{top-1} & \textit{mAP}  & \textit{top-1} & \textit{mAP}  \\
      \hline
      MNV3 & ML + Softmax & 89.03 & 90.59 & 90.92 & 91.64 \\
      MNV3 & ML + AM-Softmax, $s = \max(\sqrt{2} \cdot \log(C - 1), 3)$ & 89.082 & 92.24 & 91.35 & 93.98 \\
      MNV3 & ML + AM-Softmax, $s=1$ & \textbf{89.18} & \textbf{92.42} & \textbf{91.96} &	\textbf{94.23} \\
      \hline
    \end{tabular}
  \end{table}

\end{document}